\documentclass[11pt,a4paper]{article}
\usepackage[hyperref]{emnlp-ijcnlp-2019}
\usepackage{times}
\usepackage{latexsym}
\usepackage{graphicx}
\usepackage{float}
\usepackage{multirow}
\usepackage{url}
\usepackage{amsmath}

\aclfinalcopy 

\title{The Role of Pragmatic and Discourse Context\\ in Determining Argument Impact}

\author{Esin Durmus \\
  Cornell University \\
  \texttt{ed459@cornell.edu} \\\And
  Faisal Ladhak \\
  Amazon\\
  \texttt{faisall@amazon.com} \\\And
  Claire Cardie \\
  Cornell University \\
  \texttt{cardie@cs.cornell.edu}
  }

\date{}

\begin{document}
\maketitle
\begin{abstract}
Research in the social sciences and psychology has shown that the persuasiveness of an argument depends not only the language employed, but 
also on attributes of the source/communicator, the audience, and the appropriateness and strength of the argument's claims given the pragmatic and discourse context of the argument.
Among these characteristics of persuasive arguments, 
prior work in NLP does not explicitly investigate the effect of the pragmatic and discourse context when determining argument quality.  This paper presents a new dataset to initiate the study of this aspect of argumentation:  it consists of a diverse collection of arguments covering 741 controversial topics and
comprising over 47,000 claims.
We further propose predictive models that incorporate the pragmatic and discourse context of argumentative claims and show that they outperform models that rely only on claim-specific linguistic features for predicting the perceived impact of individual claims within a particular line of argument.


\end{abstract}
\section{Introduction}

Previous work in the social sciences and psychology has shown that the impact and
persuasive power of an argument depends not only on the language employed, but also
on the credibility and character of the communicator (i.e.\ ethos)  \cite{fb566a52435647fcbb369ed48db6fbec, Chaiken1979CommunicatorPA, source-effect}; the traits and prior beliefs of the audience \cite{audience-effect-1, audience-effect, Correll2004AnAS, doi:10.1177/0093650205277317};
and the pragmatic context in which the argument is presented (i.e.\ kairos) \cite{10.1086/209393, article-3}.
%
   
Research in Natural Language Processing (NLP) has only
partially corroborated these findings. One very influential line of work, 
for example, develops computational methods to automatically determine
the linguistic characteristics of persuasive
arguments \cite{habernal-gurevych-2016-makes, tan2016winning, zhang-etal-2016-conversational}, but it does so without controlling for
the audience, the communicator or the pragmatic context.

Very recent work, on the other hand, shows that attributes of both the 
audience and the communicator constitute important cues for determining 
argument strength \cite{lukin-etal-2017-argument,durmus-cardie-2018-exploring}.
They further show that audience and communicator attributes can influence the
relative importance of linguistic features for predicting the persuasiveness 
of an argument.
These results confirm previous findings in the social sciences that show a person's
perception of an argument can be influenced by his background and
personality traits.  

To the best of our knowledge, however, no NLP studies explicitly investigate the
role of \textit{kairos} --- a component of pragmatic context that refers to
the context-dependent ``timeliness" and ``appropriateness" of an argument and 
its claims within an argumentative discourse ---  in argument quality prediction.  
Among the many social science studies of attitude change, the order in which
argumentative claims are shared with the audience has been studied extensively:
\newcite{10.1086/209393}, for example, summarize studies showing that the
argument-related claims a person is exposed to beforehand can affect his perception
of an alternative argument in complex ways.
\newcite{article-3} similarly find that changes in an argument's context can have a big impact on the audience's perception of the argument. 

Some recent studies in NLP have investigated the effect of interactions on the overall persuasive power of posts in social media \cite{tan2016winning,AAAI1817077}. However, in social media not all posts have to express arguments or stay on topic \cite{DBLP:journals/corr/abs-1709-03167}, and qualitative evaluation of the posts can be influenced by many other factors such as interactions between the individuals \cite{Durmus:2019:MFU:3308558.3313676}. Therefore, it is difficult to measure the effect of argumentative pragmatic context alone in argument quality prediction without the effect of these confounding factors using the datasets and models currently available in this line of research.

   
\textit{In this paper, we study the role of kairos on argument quality prediction by
examining the individual claims of an argument for their timeliness and appropriateness in the context of a particular line of argument.} We define kairos as the sequence of \textbf{argumentative} text (e.g. claims) along a particular line of argumentative reasoning.

To start, we present a dataset extracted from \textit{kialo.com} of over 47,000
claims that are part of a diverse collection of arguments on 741 controversial
topics. The structure of the website dictates that each argument must present a supporting or opposing claim for its parent claim, and stay within the topic of the main thesis. Rather than being posts on a social media platform, these are community-curated claims. Furthermore, for each presented claim, the audience votes on its impact within the given line of reasoning. Critically then, the dataset includes the argument context for each claim, allowing us to
investigate the characteristics associated with impactful arguments.

With the dataset in hand, we propose the task of studying the characteristics of impactful claims by (1) taking the argument context into account, (2) studying the extent to which this context is important, and (3) determining the representation of context that is more effective. To the best of our knowledge, ours is the first dataset that includes claims with both impact votes and the corresponding context of the argument. 

\begin{figure*}[t]
\includegraphics[width=15.9cm,height=9.7cm]{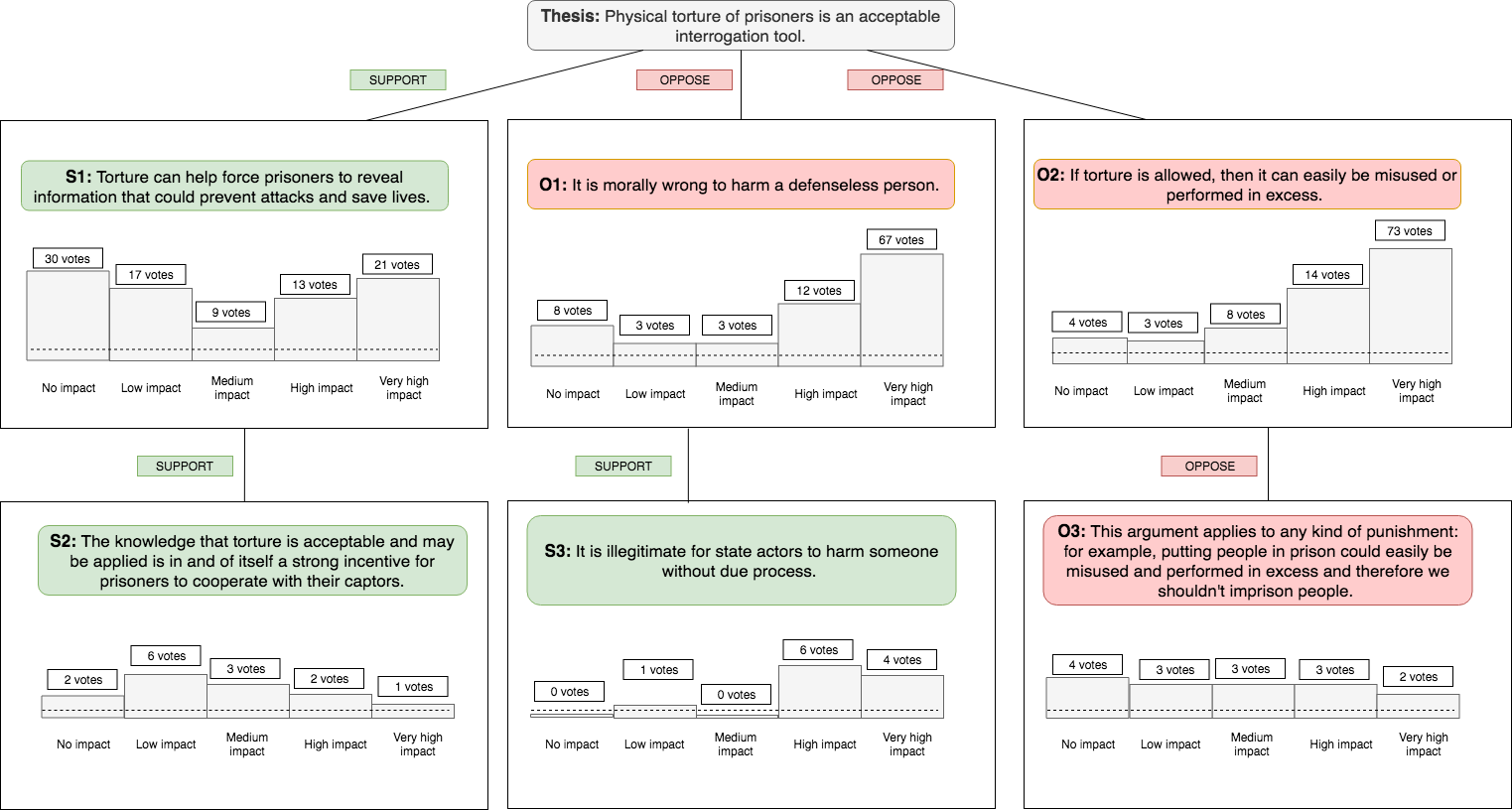}
\caption{Example partial argument tree with claims and corresponding impact votes for the thesis {\sc``Physical torture of prisoners is an acceptable interrogation tool.''}.}
\label{fig:impact_image}
\end{figure*}   
\section{Related Work}
Recent studies in computational argumentation have mainly focused on the tasks of identifying the structure of the arguments such as argument structure parsing \cite{peldszus-stede-2015-joint,park-cardie:2014:W14-21}, and argument component classification \cite{habernal-gurevych-2017-argumentation,mochales2011argumentation}. More recently, there is an increased research interest to develop computational methods that can automatically evaluate qualitative characteristic of arguments, such as their impact and persuasive power \cite{habernal-gurevych-2016-makes,tan2016winning,kelman1961processes,burgoon1975toward, chaiken1987heuristic,tykocinskl1994message,
 dillard2002persuasion, Cialdini.2007, durik2008effects, Marquart2016}.
Consistent with findings in the social sciences and psychology, some of the work in NLP has shown that the impact and persuasive power of the arguments are not simply related to the linguistic characteristics of the language, but also on characteristics the source (ethos) \cite{Durmus:2019:MFU:3308558.3313676} and the audience \cite{lukin-etal-2017-argument,durmus-cardie-2018-exploring}. These studies suggest that perception of the arguments can be influenced by the credibility of the source, and the background of the audience. 

It has also been shown, in social science studies, that \textit{kairos}, which refers to the ``timeliness'' and ``appropropriateness'' of arguments and claims, is important to consider in studies of argument impact and persuasiveness \cite{10.1086/209393,article-3}. One recent study in NLP has investigated the role of argument sequencing in argument persuasion looking at \cite{AAAI1817077} Change My View\footnote{https://www.reddit.com/r/changemyview/.}, which is a social media platform where users post their views, and challenge other users to present arguments in an attempt to change their them. However, as stated in \cite{DBLP:journals/corr/abs-1709-03167} many posts on social media platforms either do not express an argument, or diverge from the main topic of conversation. Therefore, it is difficult to measure the effect of pragmatic context in argument impact and persuasion, without confounding factors from using noisy social media data. In contrast, we provide a dataset of claims along with their structured argument path, which only consists of claims and corresponds to a particular line of reasoning for the given controversial topic. This structure enables us to study the characteristics of impactful claims, accounting for the effect of the pragmatic context. 

Consistent with previous findings in the social sciences, we find that incorporating pragmatic and discourse context is important in computational studies of persuasion, as predictive models that with the context representation outperform models that only incorporate claim-specific linguistic features, in predicting the impact of a claim. Such a system that can predict the impact of a claim given an argumentative discourse, for example, could potentially be employed by argument retrieval and generation models which aims to pick or generate the most appropriate possible claim given the discourse. 
\section{Dataset}
\textbf{Claims and impact votes.} We collected 47,219 claims from \textit{kialo.com}\footnote{The data is collected from this website in accordance with the terms and conditions.}\footnote{There is prior work by \newcite{durmus-etal-2019-determining} which created a dataset of argument trees from \textit{kialo.com}. That dataset, however, does not include any impact labels.} for 741 controversial topics and their corresponding impact votes. Impact votes are provided by the users of the platform to evaluate how impactful a particular claim is. Users can pick one of $5$ possible impact labels for a particular claim: {\sc no impact}, {\sc low impact}, {\sc medium impact}, {\sc high impact} and {\sc very high impact}. While evaluating the impact of a claim, users have access to the full argument context and therefore, they can assess how impactful a claim is in the given context of an argument. An interesting observation is that, in this dataset, the same claim can have different impact labels depending on the context in which it is presented. 

Figure \ref{fig:impact_image} shows a partial\textbf{ argument tree} for the argument \textbf{thesis} {\sc``Physical torture of prisoners is an acceptable interrogation tool.''}. Each node in the argument tree corresponds to a claim, and these argument trees are constructed and edited collaboratively by the users of the platform. 

Except the thesis, every claim in the argument tree either opposes or supports its parent claim. Each path from the root to leaf nodes corresponds to an \textbf{argument path} which represents a particular line of reasoning on the given controversial topic. 

Moreover, each claim has \textbf{impact votes} assigned by the users of the platform. The impact votes evaluate how impactful a claim is within its context, which consists of its predecessor claims from the thesis of the tree. 
For example, claim \textbf{O1} {\sc``It is morally wrong to harm a defenseless person''} is an opposing claim for the thesis and it is an {\sc impactful claim} since most of its impact votes belong to the category of {\sc very high impact}. However, claim \textbf{S3} {\sc``It is illegitimate for state actors to harm someone without the process''} is a supporting claim for its parent \textbf{O1} and it is a less impactful claim since most of the impact votes belong to the {\sc no impact} and {\sc low impact} categories. 

\begin{table}[h]
    \centering
    \begin{tabular}{|c|c|}
        \hline
       \# impact votes  & \# claims  \\
       \hline
       \hline
        $[3,5)$ & 4,495\\
        \hline
        $[5,10)$  & 5,405 \\
        \hline
        $[10,15)$  &  5,338\\
        \hline
        $[15,20)$ & 2,093\\
        \hline
        $[20,25)$ & 934\\
        \hline
        $[25,50)$ & 992\\
        \hline
        $[50,333)$ & 255\\
    
        \hline
    \end{tabular}
    \caption{Number of claims for the given range of number of votes. There are 19,512 claims in the dataset with $3$ or more votes. Out of the claims with $3$ or more votes, majority of them have $5$ or more votes.}
    \label{tab:vote_statistics_1}
\end{table}

\begin{table*}
    \centering
    \begin{tabular}{|c|c|c|}
        \hline
       & 3-class case & 5-class case \\
       \hline
       Agreement score  & Number of claims & Number of claims \\
       \hline
       \hline
       $>50\%$  & 10,848 & 7,304\\ 
       \hline
       $>60\%$ & 7,386 & 4,329\\ 
       \hline
       $>70\%$ & 4,412 & 2,195\\ 
       \hline
       $>80\%$ & 2,068 & 840\\ 
       \hline
   
    \end{tabular}
    \caption{Number of claims, with at least 5 votes, above the given threshold of agreement percentage for 3-class and 5-class cases. When we combine the low impact and high impact classes, there are more claims with high agreement score.}
    \label{tab:agreement_vote}
\end{table*}

\begin{table}[h]
    \centering
    \begin{tabular}{|c|c|}
        \hline
       Impact label  & \# votes- all claims \\
       \hline
       \hline
       No impact &  32,681\\ 
       \hline
       Low impact &  37,457\\ 
       \hline
       Medium impact &  60,136\\ 
       \hline
       High impact & 52,764\\ 
       \hline
       Very high impact & 58,846\\ 
       \hline
       Total \# votes & 241,884 \\ 
       \hline
    \end{tabular}
    \caption{Number of votes for the given impact label. There are $241,884$ total votes and majority of them belongs to the category {\sc medium impact}.  }
    \label{tab:impact_label_stats}
\end{table}
\textbf{Distribution of impact votes.} The distribution of claims with the given range of number of impact votes are shown in Table \ref{tab:vote_statistics_1}. There are 19,512 claims in total with $3$ or more votes. Out of the claims with $3$ or more votes, majority of them have $5$ or more votes.
We limit our study to the claims with at least $5$ votes to have a more reliable assignment for the accumulated impact label for each claim. 

\begin{table}
    \centering
    \begin{tabular}{|c|c|}
        \hline
       Context length  &  \# claims \\
       \hline
       \hline
       $1$ &  1,524 \\ 
       \hline
       $2$ & 1,977 \\ 
       \hline
       $3$ & 1,181\\ 
       \hline
       $[4,5]$ & 1,436 \\ 
       \hline
       $(5,10]$ & 1,115 \\ 
       \hline
       $>10$ & 153\\ 
       
       \hline
    \end{tabular}
    \caption{Number of claims for the given range of context length, for claims with more than $5$ votes and an agreement score greater than $60\%$.}
    \label{tab:context_length}
\end{table}

\textbf{Impact label statistics.} Table \ref{tab:impact_label_stats} shows the distribution of the number of votes for each of the impact categories. The claims have $241,884$ total votes. The majority of the impact votes belong to {\sc medium impact} category. We observe that users assign more {\sc high impact} and {\sc very high impact} votes than {\sc low impact} and {\sc no impact} votes respectively. When we restrict the claims to the ones with at least $5$ impact votes, we have $213,277$ votes in total\footnote{26,998 of them {\sc no impact}, 33,789 of them {\sc low impact}, 55,616 of them {\sc medium impact}, 47,494 of them {\sc high impact} and 49,380 of them {\sc very high impact.}}.

\textbf{Agreement for the impact votes.} To determine the agreement in assigning the impact label for a particular claim, for each claim, we compute the percentage of the votes that are the same as the majority impact vote for that claim. 
Let $c_{i}$ denote the count of the claims with the class labels C=[{\sc no impact}, {\sc low impact}, {\sc medium impact}, {\sc high impact}, {\sc very high impact}] for the impact label $l$ at index $i$.

\begin{equation} \label{agreement}
\text{Agreement} = 100 * \frac{\max_{0 \leq i \leq 4}c_i}{\sum_{i=0}^{4} c_i}\%\\
\end{equation}

For example, for claim S1 in Figure \ref{fig:impact_image}, the agreement score is $100 * \frac{30}{90}\%=33.33\%$ since the majority class ({\sc no impact}) has $30$ votes and there are $90$ impact votes in total for this particular claim. We compute the agreement score for the cases where (1) we treat each impact label separately (5-class case) and (2) we combine the classes  {\sc high impact} and {\sc very high impact} into a one class: {\sc impactful} and {\sc no impact} and {\sc low impact} into a one class: {\sc not impactful} (3-class case). 

Table \ref{tab:agreement_vote} shows the number of claims with the given agreement score thresholds when we include the claims with at least $5$ votes. We see that when we combine the low impact and high impact classes, there are more claims with high agreement score. This may imply that distinguishing between no impact-low impact and high impact-very high impact classes is difficult. To decrease the sparsity issue, in our experiments, we use 3-class representation for the impact labels. Moreover, to have a more reliable assignment of impact labels, we consider only the claims with have more than 60\% agreement. 

\textbf{Context.} In an argument tree, the claims from the thesis node (root) to each leaf node, form an argument path. This argument path represents a particular line of reasoning for the given thesis. Similarly, for each claim, all the claims along the path from the thesis to the claim, represent the \textbf{context} for the claim. For example, in Figure \ref{fig:impact_image}, the context for \textbf{O1} consists of only the thesis, whereas the context for \textbf{S3} consists of both the thesis and \textbf{O1} since \textbf{S3} is provided to support the claim \textbf{O1} which is an opposing claim for the thesis. 

The claims are not constructed independently from their context since they are written in consideration with the line of reasoning so far. In most cases, each claim elaborates on the point made by its parent and presents cases to support or oppose the parent claim's points. Similarly, when users evaluate the impact of a claim, they consider if the claim is timely and appropriate given its context. 
There are cases in the dataset where the same claim has different impact labels, when presented within a different context. Therefore, we claim that it is not sufficient to only study the linguistic characteristic of a claim to determine its impact, but it is also necessary to consider its context in determining the impact. 


\textit{Context length} ($\text{C}_{l}$) for a particular claim \textit{C} is defined by number of claims included in the argument path starting from the thesis until the claim \textit{C}. For example, in Figure \ref{fig:impact_image}, the context length for \textbf{O1} and \textbf{S3} are $1$ and $2$ respectively. Table \ref{tab:context_length} shows number of claims with the given range of context length for the claims with more than $5$ votes and $60\%$ agreement score. We observe that more than half of these claims have $3$ or higher context length. 

\section{Methodology}

\subsection{Hypothesis and Task Description}
Similar to prior work, our aim is to understand the characteristics of impactful claims in argumentation. However, we \textbf{hypothesize} that the qualitative characteristics of arguments is not independent of the context in which they are presented. To understand the relationship between argument context and the impact of a claim, we aim to incorporate the context along with the claim itself in our predictive models.  

\textbf{Prediction task.} Given a claim, we want to predict the impact label that is assigned to it by the users: {\sc not impactful}, {\sc medium impact}, or {\sc impactful}.

\textbf{Preprocessing.} We restrict our study to claims with at least $5$ or more votes and greater than $60\%$ agreement, to have a reliable impact label assignment. We have $7,386$ claims in the dataset satisfying these constraints\footnote{We have 1,633 {\sc not impactful}, 1,445 {\sc medium impact} and 4,308 {\sc impacful} claims.}. We see that the impact class {\sc impacful} is the majority class since around $58\%$ of the claims belong to this category.

For our experiments, we split our data to train (70\%), validation (15\%) and test (15\%) sets.

\subsection{Baseline Models}

\subsubsection{Majority} 
The majority baseline assigns the most common label of the training examples ({\sc high impact}) to every test example.

\subsubsection{SVM with RBF kernel}  
Similar to \cite{habernal-gurevych-2016-makes}, we experiment with SVM with RBF kernel, with features that represent (1) the simple characteristics of the argument tree and (2) the linguistic characteristics of the claim. 

The features that represent the simple characteristics of the claim's argument tree include the distance and similarity of the claim to the thesis, the similarity of a claim with its parent, and the impact votes of the claim's parent claim. We encode the similarity of a claim to its parent and the thesis claim with the cosine similarity of their tf-idf vectors. The distance and similarity metrics aim to model whether claims which are more similar (i.e.\ potentially more topically relevant) to their parent claim or the thesis claim, are more impactful. 

We encode the quality of the parent claim as the number of votes for each impact class, and incorporate it as a feature to understand if it is more likely for a claim to impactful given an impactful parent claim. 

\textbf{Linguistic features}. To represent each claim, we extracted the linguistic features proposed by \cite{habernal-gurevych-2016-makes} such as tf-idf scores for unigrams and bigrams, ratio of quotation marks, exclamation marks, modal verbs, stop words,  type-token ratio, hedging \cite{:/content/books/9789027282583}, named entity types, POS n-grams, sentiment \cite{ICWSM148109} and subjectivity scores \cite{wilson2005recognizing}, spell-checking, readibility features such as \textit{Coleman-Liau} \cite{1975-22007-00119750401}, \textit{Flesch} \cite{1949-01274-00119480601}, argument lexicon features \cite{somasundaran2007detecting} and surface features such as word lengths, sentence lengths, word types, and number of complex words\footnote{
We pick the parameters for SVM model according to the performance validation split, and report the results on the test split.}.

\subsubsection{FastText}
\newcite{joulin-etal-2017-bag} introduced a simple, yet effective baseline for text classification, which they show to be competitive with deep learning classifiers in terms of accuracy. Their method represents a sequence of text as a bag of n-grams, and each n-gram is passed through a look-up table to get its dense vector representation. The overall sequence representation is simply an average over the dense representations of the bag of n-grams, and is fed into a linear classifier to predict the label. We use the code released by \newcite{joulin-etal-2017-bag} to train a classifier for argument impact prediction, based on the claim text\footnote{We used maxNgram legnth of 2, learning rate of 0.8, num epochs of 15, vector dim of 300. We also used the pre-trained 300-dim wiki-news vectors made available on the fastText website.}.

\subsubsection{BiLSTM with Attention}
Another effective baseline \cite{zhou-etal-2016-attention, yang-etal-2016-hierarchical} for text classification consists of encoding the text sequence using a bidirectional Long Short Term Memory (LSTM) \cite{Hochreiter:1997:LSM:1246443.1246450}, to get the token representations in context, and then attending \cite{luong-etal-2015-effective} over the tokens to get the sequence representation. For the query vector for attention, we use a learned context vector, similar to \newcite{yang-etal-2016-hierarchical}. We picked our hyperparameters based on performance on the validation set, and report our results for the best set of hyperparameters\footnote{Our final hyperparams were: 100-dim word embedding, 100-dim context vector, 1 layer BiLSTM with 64 units, trained for 40 epochs with early stopping based on validation performance.}. We initialized our word embeddings with glove vectors \cite{pennington-etal-2014-glove} pre-trained on Wikipedia + Gigaword, and used the Adam optimizer \cite{DBLP:journals/corr/KingmaB14} with its default settings.
\begin{table*}
\centering
\begin{tabular}{|l|c|c|c|}
\hline
& Precision & Recall & F1  \\
\hline
Majority  & $19.43$ & $33.33$ & $24.55$\\
\hline 
\hline
SVM with RBF Kernel  &&&\\
\hline 
Distance from the thesis & $27.42$ & $33.53$ & $26.05$\\
\hline 
Parent quality  & $58.11$ & $47.85$ & $46.61$ \\
\hline
Linguistic features  & $\mathbf{65.67}$ & $38.58$ & $35.42$\\ 
\hline 
BiLSTM with Attention& $46.50_{\pm{0.28}}$ &$46.35_{\pm{0.99}}$& $46.22_{\pm{0.58}}$\\ 
\hline
FastText &  $51.18_{\pm{0.80}}$ &$46.09_{\pm{0.64}}$& $47.06_{\pm{0.70}}$\\ 
\hline
BERT models & & & \\
\hline
Claim only  & $53.24_{\pm{1.07}}$ &$50.93_{\pm{2.01}}$& $51.53_{\pm{1.53}}$\\ 
\hline 
Claim + Parent  & $55.79_{\pm{1.72}}$ &  $53.54_{\pm{2.09}}$ & $54.00_{\pm{1.79}}$\\ 
\hline
Claim + $\text{Context}_{f}(2)$  & $56.57_{\pm{0.85}}$ &  $54.76_{\pm{1.71}}$ & $55.18_{\pm{0.99}}$\\
\hline
Claim + $\text{Context}_{f}(3)$ &  $57.19_{\pm{0.92}}$ & $\mathbf{55.77_{\pm{1.05}}}$ & $\mathbf{55.98_{\pm{0.70}}}$\\
\hline 
Claim + $\text{Context}_{f}(4)$ &  $57.09_{\pm{1.71}}$ & $55.31_{\pm{1.09}}$ & $55.72_{\pm{1.14}}$\\
\hline
Claim + $\text{Context}_{gru}(4)$ &  $54.95_{\pm{2.00}}$ & $51.55_{\pm{1.27}}$ & $52.37_{\pm{1.26}}$\\
\hline
Claim + $\text{Context}_{a}(4)$   & $56.60_{\pm{0.52}}$ & $54.55_{\pm{0.57}}$ & $54.65_{\pm{0.33}}$ \\ 
\hline 

\end{tabular}
\caption{Results for the baselines and the BERT models with and without the context. Best performing model is BERT with the representation of previous $3$ claims in the path along with the claim representation itself. We run the models $5$ times and we report the mean and standard deviation. }
\label{tab:results}
\end{table*}

\subsection{Fine-tuned BERT model}
\newcite{devlin2018bert} fine-tuned a pre-trained deep bi-directional transformer language model (which they call BERT), by adding a simple classification layer on top, and achieved state of the art results across a variety of NLP tasks. We employ their pre-trained language models for our task and compare it to our baseline models. For all the architectures described below, we finetune for 10 epochs, with a learning rate of 2e-5. We employ an early stopping procedure based on the model performance on a validation set.
\subsubsection{Claim with no context}
In this setting, we attempt to classify the impact of the claim, based on the text of the claim only. We follow the fine-tuning procedure for sequence classification detailed in \cite{devlin2018bert}, and input the claim text as a sequence of tokens preceded by the special [CLS] token and followed by the special [SEP] token. We add a classification layer on top of the BERT encoder, to which we pass the representation of the [CLS] token, and fine-tune this for argument impact prediction.

\subsubsection{Claim with parent representation} 
In this setting, we use the parent claim's text, in addition to the target claim text, in order to classify the impact of the target claim. We treat this as a sequence pair classification task, and combine both the target claim and parent claim as a single sequence of tokens, separated by the special separator [SEP]. We then follow the same procedure above, for fine-tuning.

\subsubsection{Incorporating larger context} 
In this setting, we consider incorporating a larger context from the discourse, in order to assess the impact of a claim. In particular, we consider up to four previous claims in the discourse (for a total context length of 5). We attempt to incorporate larger context into the BERT model in three different ways.

\textbf{Flat representation of the path.}
The first, simple approach is to represent the entire path (claim + context) as a single sequence, where each of the claims is separated by the [SEP] token. BERT was trained on sequence pairs, and therefore the pre-trained encoders only have two segment embeddings \cite{devlin2018bert}. So to fit multiple sequences into this framework, we indicate all tokens of the target claim as belonging to segment A and the tokens for all the claims in the discourse context as belonging to segment B.  This way of representing the input, requires no additional changes to the architecture or retraining, and we can just finetune in a similar manner as above. We refer to this representation of the context as a flat representation, and denote the model as $\text{Context}_{f}(i)$, where $i$ indicates the length of the context that is incorporated into the model.
\begin{table*}[h]
\centering
\begin{tabular}{|l|c|c|c|c|c|}
\hline
& $\text{C}_{l}=1$ & $\text{C}_{l}=2$ & $\text{C}_{l}=3$ & $\text{C}_{l}=4$ \\
\hline
BERT models & & & &   \\
\hline
\hline
Claim only & $48.61_{\pm{3.16}}$ & $53.15_{\pm{1.95}}$ & $54.51_{\pm{1.91}}$ &  $50.89_{\pm{2.95}}$\\ 
\hline 
Claim + Parent & $51.49_{\pm{2.63}}$ & $54.78_{\pm{2.95}}$& $54.94_{\pm{2.72}}$ & $51.94_{\pm{2.59}}$ \\
\hline
Claim +  $\text{Context}_{f}(2)$ & $52.84_{\pm{2.55}}$ & $53.77_{\pm{1.00}}$ & $55.24_{\pm{2.52}}$ &  $57.04_{\pm{1.19}}$\\ 
\hline
Claim + $\text{Context}_{f}(3)$  & $\mathbf{54.88_{\pm{2.49}}}$ &  $54.71_{\pm{1.74}}$ & $52.93_{\pm{2.07}}$& $\mathbf{58.17_{\pm{1.89}}}$ \\
\hline 
Claim + $\text{Context}_{f}(4)$ &  $54.47_{\pm{2.95}}$ & $\mathbf{54.88_{\pm{1.53}}}$ &$\mathbf{57.11_{\pm{3.38}}}$ & $57.02_{\pm{2.22}}$ \\
\hline
\end{tabular}
\caption{F1 scores of each model for the claims with various context length values.}
\label{tab:results_context}
\end{table*}

\textbf{Attention over context.}
Recent work in incorporating argument sequence in predicting persuasiveness \cite{AAAI1817077} has shown that hierarchical representations are effective in representing context. Similarly, we consider hierarchical representations for representing the discourse. 
We first encode each claim using the pre-trained BERT model as the claim encoder, and use the representation of the [CLS] token as claim representation. 
We then employ dot-product attention \cite{luong-etal-2015-effective}, to get a weighted representation for the context. We use a learned context vector as the query, for computing attention scores, similar to \newcite{yang-etal-2016-hierarchical}. The attention score $\alpha_c$ is computed as shown below:
\begin{equation}
    \alpha_{c} = \frac{exp(V_{c}^T V_{l})}{\sum_{c\in{D}} exp(V_{c}^T V_{l})}
\end{equation}

Where $V_c$ is the claim representation that was computed with the BERT encoder as described above, $V_l$ is the learned context vector that is used for computing attention scores, and $D$ is the set of claims in the discourse.
After computing the attention scores, the final context representation $v_d$ is computed as follows:
\begin{equation}
     V_{d} = \sum_{c\in{D}}\alpha_{c} V_{c}
\end{equation}
We then concatenate the context representation with the target claim representation $[V_d, V_r]$ and pass it to the classification layer to predict the quality. We denote this model as $\text{Context}_{a}(i)$.

\textbf{GRU to encode context}
Similar to the approach above, we consider a hierarchical representation for representing the context. We compute the claim representations, as detailed above, and we then feed the discourse claims' representations (in sequence) into a bidirectional Gated Recurrent Unit (GRU) \cite{cho-al-emnlp14}, to compute the context representation. We concatenate this with the target claim representation and use this to predict the claim impact. We denote this model as $\text{Context}_{gru}(i)$.



\section{Results and Analysis}
Table \ref{tab:results} shows the macro precision, recall and F1 scores for the baselines as well as the BERT models with and without context representations\footnote{For the models that result in different scores with different random seed, we run them $5$ times and report the mean and standard deviation.}. 

We see that \textit{parent quality} is a simple yet effective feature and SVM model with this feature can achieve significantly higher ($p<0.001$)\footnote{We perform two-sided t test for significance analysis.} F1 score ($46.61\%$) than \textit{distance from the thesis} and \textit{linguistic features}. Claims with higher impact parents are more likely to be have higher impact. \textit{Similarity with the parent and thesis} is not significantly better than the \textit{majority} baseline. Although the BiLSTM model with attention and FastText baselines performs better than the SVM with \textit{distance from the thesis} and \textit{linguistic features}, it has similar performance to the \textit{parent quality} baseline.

We find that the BERT model with \textit{claim only} representation performs significantly better ($p<0.001$) than the baseline models.  Incorporating the \textit{parent representation} only along with the \textit{claim representation} does not give significant improvement over representing the claim only. However, \textit{incorporating the flat representation of the larger context} along with the claim representation consistently achieves significantly better ($p<0.001$) performance than the claim representation alone. Similarly, \textit{attention representation} over the context with the learned query vector achieves significantly better performance then the \textit{claim representation} only ($p<0.05$). 

We find that the \textit{flat representation} of the context achieves the highest F1 score. It may be more difficult for the models with a larger number of parameters to perform better than the \textit{flat representation} since the dataset is small. We also observe that modeling $3$ claims on the argument path before the target claim achieves the best F1 score ($55.98\%$).

To understand for what kinds of claims the best performing contextual model is more effective, we evaluate the BERT model with \textit{flat context representation} for claims with context length values $1$, $2$, $3$ and $4$ separately. Table \ref{tab:results_context} shows the F1 score of the BERT model without context and with \textit{flat context representation} with different lengths of context. For the claims with context length $1$, adding $\text{Context}_{f}(3)$ and $\text{Context}_{f}(4)$ representation along with the claim achieves significantly better $(p<0.05)$ F1 score than modeling the \textit{claim only}. Similarly for the claims with context length $3$ and $4$, $\text{Context}_{f}(4)$ and $\text{Context}_{f}(3)$ perform significantly better than BERT with \textit{claim only} ($(p<0.05)$ and $(p<0.01)$ respectively). We see that models with larger context are helpful even for claims which have limited context (e.g. $\text{C}_{l}=1$). This may suggest that when we train the models with larger context, they learn how to represent the claims and their context better. 
\section{Conclusion}
In this paper, we present a dataset of claims with their corresponding impact votes, and investigate the role of argumentative discourse context in argument impact classification. We experiment with various models to represent the claims and their context and find that incorporating the context information gives significant improvement in predicting argument impact. In our study, we find that \textit{flat representation} of the context gives the best improvement in the performance and our analysis indicates that the contextual models perform better even for the claims with limited context.

\section{Acknowledgements}
This work was supported in part by NSF grants IIS-1815455 and SES-1741441.  The views and conclusions contained herein are those of the authors and should not be interpreted as necessarily representing the official policies or endorsements, either expressed or implied, of NSF or the U.S.\ Government.

\bibliography{emnlp-ijcnlp-2019}
\bibliographystyle{acl_natbib}
\end{document}